\pdfoutput=1

\documentclass[11pt]{article}

\usepackage{acl}

\usepackage{times}
\usepackage{latexsym}
\usepackage{enumitem}

\usepackage[T1]{fontenc}

\usepackage[utf8]{inputenc}

\usepackage[nopatch=footnote]{microtype}

\usepackage{inconsolata}

\usepackage[T1]{fontenc}
\usepackage{type1cm}
\usepackage[american]{babel}
\usepackage{url}
\usepackage{amssymb}
\usepackage{amsmath}
\usepackage{multirow}
\usepackage{xspace}
\DeclareMathAlphabet{\mathcalbf}{OMS}{pzc}{b}{n}
\usepackage{lipsum}

\usepackage[pdftex]{graphicx}
\usepackage{tabularx}
\usepackage{booktabs}
\DeclareGraphicsExtensions{.pdf,.ai,.jpg,.png}
\setkeys{Gin}{pagebox=artbox}
\usepackage{subcaption}

\usepackage{array}
\usepackage{makecell}
\usepackage{setspace}

\graphicspath{{co-constructive-llms-figures/}}

\RequirePackage{type1cm}
\RequirePackage{color}
\RequirePackage{soul}
\setstcolor{blue}
\definecolor{violet}{rgb}{0.5,0.0,0.5}
\newsavebox\bscombox
\newcommand{\bscom}[3][]{%
	\sbox{\bscombox}{\fontsize{8}{9}\selectfont#1#2#3}
	\noindent
	\st{#2}{\selectfont
		\color{blue}#3\ifx\\#1\\\else{\fontsize{8}{9}\selectfont\color{violet}[#1]}\fi
	}
}

\definecolor{highlight1}{rgb}{0.95,0.95,0.95}
\definecolor{tgray}{rgb}{0.5,0.5,0.5}
\definecolor{tgray}{rgb}{0.5,0.5,0.5}

\begin{document}

\title{Investigating Co-Constructive Behavior of\\Large Language Models in Explanation Dialogues}

\author{
	\textbf{Leandra Fichtel}\,\textsuperscript{\rm 1}\thanks{\enspace Both authors contributed equally to this paper.},
	\textbf{Maximilian Spliethöver}\,\textsuperscript{\rm 1}\footnotemark[1],
	\textbf{Eyke Hüllermeier}\,\textsuperscript{\rm 2},
	\textbf{Patricia Jimenez}\,\textsuperscript{\rm 3},\\
	\textbf{Nils Klowait}\,\textsuperscript{\rm 3},
	\textbf{Stefan Kopp}\,\textsuperscript{\rm 4},
	\textbf{Axel-Cyrille Ngonga Ngomo}\,\textsuperscript{\rm 3},
	\textbf{Amelie Robrecht}\,\textsuperscript{\rm 4},\\
	\textbf{Ingrid Scharlau}\,\textsuperscript{\rm 3},
	\textbf{Lutz Terfloth}\,\textsuperscript{\rm 3},
	\textbf{Anna-Lisa Vollmer}\,\textsuperscript{\rm 4},
	\textbf{Henning Wachsmuth}\,\textsuperscript{\rm 1} \\
	\textsuperscript{1}\,Leibniz University Hannover, Institute of Artificial Intelligence\\
	\textsuperscript{2}\,LMU Munich, MCML \quad
	\textsuperscript{3}\,Paderborn University \quad
	\textsuperscript{4}\,Bielefeld University, CITEC\\
	\texttt{\href{mailto:l.fichtel@ai.uni-hannover.de}{l.fichtel@ai.uni-hannover.de}} \\
	\texttt{\href{mailto:m.spliethoever@ai.uni-hannover.de}{m.spliethoever@ai.uni-hannover.de}}
}
\maketitle
\begin{abstract}
The ability to generate explanations that are understood by explainees is the quintessence of explainable artificial intelligence. Since understanding depends on the explainee's background and needs, recent research focused on co-constructive explanation dialogues, where an explainer continuously monitors the explainee's understanding and adapts their explanations dynamically. We investigate the ability of large language models (LLMs) to engage as explainers in co-constructive explanation dialogues. In particular, we present a user study in which explainees interact with an LLM in two settings, one of which involves the LLM being instructed to explain a topic co-constructively. We evaluate the explainees' understanding before and after the dialogue, as well as their perception of the LLMs' co-constructive behavior. Our results suggest that LLMs show some co-constructive behaviors, such as asking verification questions, that foster the explainees' engagement and can improve understanding of a topic. However, their ability to effectively monitor the current understanding and scaffold the explanations accordingly remains limited.
\end{abstract}

\section{Introduction}
\label{sec:introduction}

\begin{figure}
	\centering
	\includegraphics[scale=1]{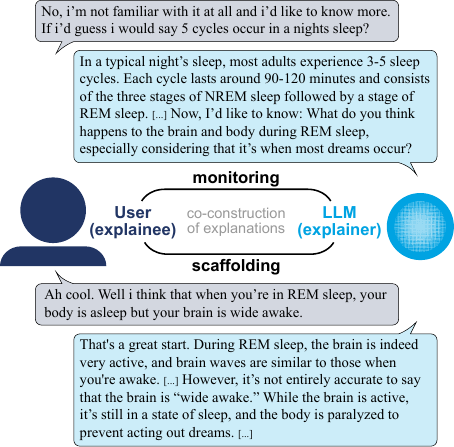}
	\caption{An explanation dialogue about the sleep cycle from our study between a user and an LLM, acting as explainee and explainer, respectively. We analyze to what extent LLMs can monitor and scaffold.}
	\label{fig:example-chat}
\end{figure}

The computational generation of natural language explanations has gained research interest due to its importance for explainable artificial intelligence (XAI), which aims to explain decisions made by AI systems to humans~\cite{schneider2019personalized}. A key challenge in XAI is the asymmetry between system and human, not only in terms of information access (e.g., only the system has access to information on internal decision processes) but also in terms of the human's knowledge and level of expertise~\cite{miller2019whathowwhy}. Therefore, to achieve understanding, it is important to account for the diverse backgrounds and individual \nopagebreak {abilities of} humans~\cite{wachsmuth2022wired}. Recent XAI research, thus, focuses on personalized explanations that aim to improve their effectiveness~\cite{Sokol2020Personalized,hostetter2023,mindlin2024,nimmo2024}. However, in real-world social interactions, understanding dynamically evolves in dialogues between explainers and explainees. Therefore, effective explanations should not only involve an initial personalization but also continuously adapt to the explainee's needs throughout the interaction \citep{robrecht2023}. This can be achieved in a \emph{co-constructive} explanation dialogue in which the explainer and explainee construct understanding interactively. For this, the explainer continuously \textit{monitors} the explainee's understanding and \textit{scaffolds} (i.e. adapts) explanations accordingly, as exemplified in Figure~\ref{fig:example-chat}~\cite{Molenaar2011scaffolding,rohlfing2020socialpractice}. The question arises as to how to enable an XAI system to lead such co-constructive explanation dialogues.

Large language models (LLMs) have made significant progress in recent years, demonstrating a remarkable ability to generate coherent and contextually relevant text in various tasks~\citep{dubey2024}. Fine-tuning LLMs to follow instructions~\cite{ouyang2022RLHF,wang2023selfinstruct} has further enhanced these capabilities, enabling LLMs to adjust their behavior to complex prompts~\cite{spliethoever:2025}, personalize the interaction experience by adopting specific personas~\citep{chen2024}, and to support users to construct knowledge~\cite{Cress2023CoconstructingKnowledge}. However, it remains unclear so far whether these capabilities also enable co-constructive explanation dialogues.

In this paper, we study co-constructive explanations with LLMs, focusing on three questions:
\begin{enumerate}[itemsep=3pt,parsep=0pt,topsep=3pt,label=\arabic*.]
    \item How to enable co-constructive explanation dialogues using LLMs as explainers?
    \item To what extent do LLMs show co-constructive behaviors?
    \item How effectively do LLMs guide explainees toward a better understanding of a given topic?
\end{enumerate}

To answer these questions, we conduct a user study in which participants interact with an LLM to receive explanations about a predefined topic. We test one LLM with two zero-shot settings based on different system prompts: In the \textit{base setting}, the LLM is simply instructed to act as an explainer. By contrast, in the \textit{enhanced setting}, the LLM is given detailed instructions to follow co-constructive behavior by applying monitoring and scaffolding.

We analyze the resulting data both quantitatively and qualitatively, focusing on the participants' understanding of the topic (\textit{comprehension}), their ability to perform actions in the domain of the topic (\textit{enabledness})~\cite{buschmeier2023formsunderstanding}, and on the co-constructive behavior of the LLM.

Our results indicate that the enhanced setting can enable the evaluated LLM to exhibit co-constructive behavior, like monitoring the explainee's understanding via verification questions and encouraging active participation. These co-constructive behaviors seem to increase explanatory success in selected cases.
However, the success in monitoring and scaffolding seems to be rather inconsistent.
In addition, the LLM tends toward monologues that leave little room for interaction in both settings.%
\footnote{The code and data can be found under \url{https://github.com/webis-de/SIGDIAL-25}.}

\section{Related Work}
\label{sec:relatedwork}

Recently, the abilities of LLMs to explain concepts, decisions, or behavior have been explored in NLP research~\cite{dibonaventura2024,kunz2024}. Many approaches focus on the generation of a single natural language explanation, partly in response to a question or another description of what is to be explained~\cite{rajani2019,fan2019}. As discussed in Section~\ref{sec:introduction}, however, there is not \emph{the} right explanation in many real-world situations, due to the different backgrounds and needs of explainees~\cite{wachsmuth2022wired}. Therefore, we look at explanation \emph{dialogues}~\cite{elassady2019XAI,rohlfing2020socialpractice} in which an explainer interacts with an explainee to co-construct an explanation.

One way to instruct LLMs to enact the role of an explainer is through \textit{persona prompting}, i.e., assigning a predefined persona. Among others, this technique has been used in NLP to diversify automated data annotations~\cite{giorgi2024, beck2024}, and in the social sciences to simulate specific samples of the human population, i.e., social groups. Instead of using persona prompts to annotate data or simulate samples of social groups, we aim to utilize personas to simulate an explainer with co-constructive behavior in a dialogue setting.

Even though instruction-tuned LLMs are often used in dialogue settings, they were originally optimized to follow instructions~\citep{ouyang2022RLHF,wang2022instruct}, with selected variants being tuned towards dialogues~\citep{ding:2023,deng:2024}. \citet{wang2024} aim to model the dialogue explicitly by modeling each conversation party separately instead of mixing them in a single context.~\citet{andukuri2024stargate}, on the other hand, develop an approach that learns to ask clarification questions, a specific dialogue act~\cite{bunt:2010}, when a user query does not contain sufficient information. Focusing more on the interaction in explanation dialogues,~\citet{wachsmuth2022wired} present a corpus to formalize dialogue acts and explanation moves common in a co-constructive setting.~\citet{alshomary:2024} use this corpus to automatically identify the respective acts and moves to estimate the quality of an explanation. In contrast, we do not model the dialogue explicitly or implement certain explanation moves. Rather, we evaluate how well out-of-the-box LLMs can be instructed to act co-constructively and to utilize the dialogue context for actions such as monitoring the explainee and scaffolding explanations.

Our notion of co-constructive LLMs is closely related to the concept of automated tutoring systems, i.e., computational systems that can teach a topic to a human student and adapt to individual needs.~\citet{forbes-riley2011} and \citet{robrecht2023} evaluate the value of such adaptation and find that it can significantly improve the learning effect. \citet{cawsey1993} and \citet{robrecht2025} propose adaptive systems that aim to dynamically decide on the best explanation strategy at any given point in the interaction with the student. In this work, we do not build a complex architecture or evaluate the effectiveness of adaptations. Rather, we aim to evaluate the capabilities of out-of-the-box LLMs to monitor the explainee implicitly and scaffold its explanations accordingly.

Related to our work,~\citet{hoffman2023} evaluate LLM-generated explanations from the explainee's perspective and present several criteria to measure success in explaining AI systems.~\citet{danry2023} further find that having the explainer ask critical questions instead of making factual statements can enhance the explanatory success. Lastly,~\citet{Klowait2024AIexplainAI} investigate whether a GPT-4-based LLM can act as an explainer in an XAI setting. They find some co-constructive patterns in the LLM's responses, but only when actively engaged as such by the participants. We, however, focus on an explanation dialogue setting from the explainer's perspective, and evaluate how much co-constructive behavior can be achieved with prompting. Furthermore, our analysis focuses on dialogue acts and moves, and the ability of the LLM to enact monitoring and scaffolding.

\section{Co-Constructive Explanations}
\label{sec:coconstruct}

An effective way to improve explanations is through a co-constructive explanation dialogue~\cite{robrecht2023}. In this section, we briefly summarize the basic concepts of co-constructive explanations, which we investigate in this study. Co-construction is considered an important micro-level aspect in a conversation between an explainer (i.e., the party who explains), and an explainee (i.e., the party who is explained to)~\cite{rohlfing2020socialpractice}. The goal of a co-constructive explanation is to co-construct \textit{understanding} of the topic being explained, the so-called \emph{explanandum}~\cite{hempel1948explanandum,Lombrozo2006explanandum}. This can be achieved by both parties \emph{monitoring} each other's understanding and \emph{scaffolding} explanations accordingly~\cite{Molenaar2011scaffolding,rohlfing2020socialpractice}. Below, we detail the central concepts of co-constructive processes which are relevant to our evaluation. Since we evaluate the co-constructive behavior of LLMs acting as explainers, we present the concepts from the explainer's perspective.

\paragraph{Explanandum}

The explanandum is the subject of an explanation dialogue, which is explained to the explainee~\cite{hempel1948explanandum,Lombrozo2006explanandum}. While there is usually an initial explanandum from which a co-constructive explanation dialogue starts, the explanandum is adapted by both parties throughout the dialogue~\cite{rohlfing2020socialpractice,Booshehri2024coconstruct}.

\paragraph{Monitoring}

A crucial part of effective explanations involves monitoring the explainee through diagnostic and verification questions to identify their knowledge gaps~\cite{elassady2019XAI}. Commonly, a knowledge gap is assumed to be identifiable and static before a dialogue. However,~\citet{rohlfing2020socialpractice} argue that identifying and agreeing on the knowledge gap emerges from the dialogue. The explainer, thus, has to continuously assess and verify the explainee's understanding to establish a common ground and define the explanandum.

\paragraph{Scaffolding}

Based on the results of the monitoring, scaffolding involves the explainer adjusting the level of assistance and adapting the explanations according to the explainee's current understanding~\cite{Molenaar2011scaffolding,rohlfing2020socialpractice}. The goal is to focus on aspects of the given explanandum that are within the abilities of the explainee~\cite{rohlfing2020socialpractice}.

\paragraph{Understanding}

One measure of explanatory success is the extent to which the explainee's understanding of the explanandum improves.
\citet{buschmeier2023formsunderstanding} define understanding as a combination of \textit{comprehension}, also known as conceptual knowledge, and \textit{enabledness}, i.e., the explainee's ability to perform specific actions in the context of the explanandum.
We further distinguish between subjective and objective understanding. Whereas subjective understanding describes the explainees' self-assessed understanding of the explanandum, objective understanding measures their actual understanding~\cite{buschmeier2023formsunderstanding}.

\section{Co-Constructive LLMs}
\label{sec:cocon-llms}

Conceptually, the introduced process of co-constructing explanations appears to fit well the interaction capabilities of instruction-tuned LLMs: Although such LLMs are not inherently designed for the role of a co-constructive explainer, their pretraining and fine-tuning allow them to follow concrete instructions in a dialogical setting~\cite{ouyang2022RLHF}. Therefore, it might also be possible to enable co-constructive behavior in explanation dialogues by instructing such LLMs accordingly.

We thus explore whether instruction-tuned LLMs can effectively lead co-constructive explanation dialogues. In particular, we assess the co-constructive potential of existing LLMs that is accessible to all users through a \textit{system prompt}, without the need for complex tuning. We evaluate two settings that resemble different levels of prompt complexity, a \textit{base setting} and an \textit{enhanced setting}. Figure~\ref{fig:prompts-short} presents the prompt of each setting (see Appendix~\ref{apdx:full-prompts} for further details).

\paragraph{Base Setting}

In the base setting, the prompt provides only basic information about the scenario; namely, it instructs the LLM to act as an explainer. This setting is used to assess how LLMs behave by default and assess the impact of a detailed prompt.

\paragraph{Enhanced Setting}

The enhanced setting uses a prompt that provides more details about the co-constructive explanation setting. Specifically, it instructs the LLM to act as a co-constructive explainer by applying monitoring and scaffolding, and provides definitions of these two core concepts.

\begin{figure}
    \centering
    \includegraphics[scale=1]{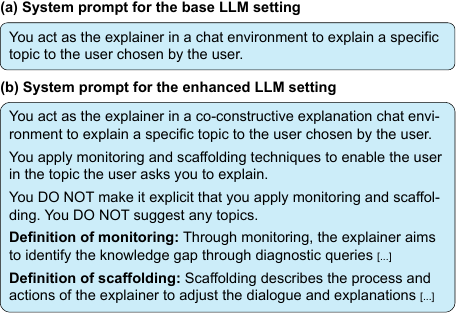}
    \caption{System prompts used for the LLM in the \textit{base} and \textit{enhanced} settings. They include different instructions related to the LLM’s desired behavior.}
    \label{fig:prompts-short}
\end{figure}

\medskip
We expect that the prompts can be further optimized towards the co-constructive setting, which could result in better co-constructive behavior of the LLM. However, this may require a rigorous evaluation of different prompt settings~\cite{spliethoever:2025}. Since our focus is the co-constructive behavior of LLMs in general rather than optimizing the system prompt, we leave this to future work.

\section{Experimental Setup}
\label{sec:user-study}

To evaluate the co-constructive behavior of LLMs in explanation dialogues, we conduct a user study in which participants receive explanations about one of three \emph{explananda} by interacting with an LLM acting as an explainer.
The participants complete pre- and post-interaction \emph{questionnaires} to assess their understanding of the explanandum, their motivation to learn about it, and their perceptions of the LLM's co-constructive behavior.

The user study was conducted on Prolific. We hired participants who are fluent in English and have completed at least 500 submissions with an approval rate of at least 95\%.

\subsection{Chat Application}
\label{sec:chat-application}
Our user study application (see Appendix~\ref{apdx:study-application} for details) guides the participants through a pre-interaction questionnaire, an interface to interact with the LLM, and several post-interaction questionnaires.

\paragraph{LLM Selection}

We use the instruction-tuned variant of the open-weight model Llama 3.1~\cite{dubey2024} with 70 billion parameters (details in Appendix~\ref{apdx:hyperparamters}). While larger open-weight models are available, we prioritize timely responses to ensure a natural interaction experience. We avoid closed-weight LLMs as such models change frequently and may not produce consistent results over time. We further set a fixed seed to ensure reproducibility.

\paragraph{LLM Interaction}

Each participant interacts with the LLM in either the \textit{base setting} or the \textit{enhanced setting} described above. To provide a clear goal, we instruct the participants to learn as much as possible about the given initial explanandum through the interaction until they feel confident in explaining it to someone.
Based on the findings of a pre-study, we decide against priming the participants with details about their interaction partner (details in Appendix~\ref{apdx:pre-study}). For consistency, we limit the interaction time to 15 minutes per participant.

\subsection{Initial Explananda}
\label{sc:initial-explananda}
We provide the participants with an initial explanandum to frame the explanation dialogue and reduce the potential for deviations from the evaluated explanandum by unifying their starting point.

We evaluate three diverse explananda to improve the reliability and generalizability of our findings:
\begin{enumerate}[itemsep=0pt,parsep=0pt,topsep=3pt,label=\arabic*.]
    \item The board game \textit{Quarto} and its rules
    \item The formation of \textit{black holes}
    \item The human \textit{sleep} cycle and its stages
\end{enumerate}

\noindent
To limit the influence of extraneous variables on the explanation dialogues, we select the initial explananda to be understandable without extensive background knowledge but still complex enough to prevent complete understanding. They should be common enough for LLMs to generate plausible explanations, and broadly relevant to avoid demographic bias. See Appendix~\ref{apdx:global-explanada-selection} for details.

\subsection{Questionnaires}
\label{sec:questionnaires}

The participants complete a total of five questionnaires that assess explanatory success through their understanding of the initial explanandum and provide insights into the LLM's co-constructive behavior. All full questionnaires are in Appendix~\ref{apdx:questionniare-details}.

\paragraph{Comprehension and Enabledness}

Four questionnaires assess the participants' understanding of the initial explanandum in terms of comprehension and enabledness~\cite{buschmeier2023formsunderstanding}, one before and three after the LLM interaction.

Before interacting with the LLM, the first questionnaire asks about the pre-existing \emph{subjective comprehension} of the initial explanandum~\cite{buhl2025partnermodelquestionnaire}, and the \emph{motivation} regarding the initial explanandum~\cite{Rheinberg2001motivationquestionnaire,strecker2002motivationquestionnaire,Schiefele2016motivationquestionnaire,buhl2025partnermodelquestionnaire}, both assessed on a five-point Likert scale. After the interaction, we present a slightly modified version of the questionnaire again to capture potential changes.

In addition, after the interaction, the participants complete two questionnaires derived from~\citet{terfloth2025} to assess the participants' objective understanding in terms of \emph{objective comprehension} and their \emph{enabledness}. The objective comprehension questionnaire consists of 14 agree/disagree statements. The enabledness questionnaire contains five multiple-choice questions. To prevent learning effects, we test for objective understanding only after the interaction.

\paragraph{Co-Constructive Behavior}
Finally, the participants complete a post-interaction questionnaire to assess potential co-constructive behaviors of the LLM, including monitoring and scaffolding. Because our LLM-participant setting is similar to a teacher-student setting, for our questionnaire, we rely on the scales manual of \citet{buhl2025} which utilizes items of the \textit{Approaches to Teaching Inventory} (ATI)~\cite{staub2002teacher,trigwell2004teacher,roscoe2008teacher,roscoe2014teacher}. The ATI is designed to help teachers understand how their strategies affect student learning. Among others, \citet{buhl2025} adopt items of the ATI that are related to co-constructive behavior, e.g., \textit{While explaining, it was important to my dialogue partner to continuously consider if I understood the explanation.} The participants rate the statements on a five-point Likert scale.

\subsection{Evaluation Measures}

Focusing on the aspects introduced in Section~\ref{sec:coconstruct}, we evaluate the LLMs' co-constructive behavior intrinsically and extrinsically based on the dialogues and questionnaires. For detailed evaluations, we measure significant differences with the Mann-Whitney U Test~\cite{mann1947mannWhitneyUTest} and correlations in terms of Kendall's $\tau$~\cite{kendall1938tau}.

\paragraph{Intrinsic Evaluation}

Our intrinsic evaluation assesses the LLM's co-constructive behavior. For this, we automatically annotate the explanation moves and dialogue acts of the turns within the dialogues, using the approach of~\citet{alshomary:2024} . See Appendix~\ref{apdx:turn-level-prediction-setup} for details. In addition, noting that this is only one possible scaffolding technique, we assess the readability of the LLMs' explanations using the Gunning Fog Index~\citep{gunning1968}, Type-Token Ratio~\citep{johnson1944}, and Shannon Entropy~\citep{shi2022,shannon1948entropy}. Lastly, we conduct a qualitative analysis on selected dialogues to gain deeper insights into the LLM's co-constructive behavior.

\paragraph{Extrinsic Evaluation}
With the extrinsic evaluation, we measure explanatory success through a quantitative analysis of the understanding questionnaires.
In addition, we evaluate the participants' engagement using quantitative metrics, such as the number of queries initiated by the participants or the processing time taken to read the LLM's responses and to formulate their new query.

\section{Results and Discussion}
\label{sec:results}

In total, 300 participants completed the study. To ensure high-quality data, we excluded from the study those who did not ask questions about their initial explanandum. After filtering, 277 participants in total and at least 45 per LLM setting remain (see Appendix~\ref{apdx:study-statistics} for exact counts per topic). Below, we first evaluate the co-constructive behavior in both LLM settings, followed by the participants' understanding to assess explanatory success.

\subsection{Engagement of Participants}
\label{sec:engagement-results}

\begin{table}[t]
    \small
    \centering

    \renewcommand{\arraystretch}{0.95}
    \setlength{\tabcolsep}{1.5pt}

    \begin{tabular}{l@{\,}rrrrr}
        \toprule

        & & \multicolumn{2}{c}{\bf Explainee} & \multicolumn{2}{c}{\bf Explainer} \\
        \cmidrule(l@{3pt}r@{3pt}){3-4}\cmidrule(l@{3pt}r@{3pt}){5-6}

        \bf Setting & \bf Duration &  \bf Queries & \bf Process. & \bf Sent's & \bf Words \\

        \midrule

        Base 	& \textsuperscript{$\dagger\!$}12:26 {\tiny $\!\pm$216s} 	& \textsuperscript{$\dagger\!$}8.2 {\tiny $\!\pm$3.8} 	& 01:49 {\tiny $\!\pm$64s} 	& \textsuperscript{$\dagger\!$}17.8 {\tiny $\!\pm$5.3} 	& \textsuperscript{$\dagger\!$}16.8 {\tiny $\!\pm$2.6} \\
        Enhanc. 	& \textsuperscript{$\dagger\!$}13:25 {\tiny $\!\pm$175s} 	& \textsuperscript{$\dagger\!$}9.3 {\tiny $\!\pm$3.7} 	& 01:40 {\tiny $\!\pm$51s} 	& \textsuperscript{$\dagger\!$}12.0 {\tiny $\!\pm$3.6} 	& \textsuperscript{$\dagger\!$}17.8 {\tiny $\!\pm$2.2} \\

        \bottomrule
    \end{tabular}

    \caption{Dialogue statistics for the \emph{base} and \emph{enhanced} settings, averaged across all topics, showing the \textit{duration} (min), the number of \textit{queries} and the \textit{processing time} (min) for the participants (explainee), and the average numbers of \textit{sentences} and \textit{words per sentence} per LLM response (explainer). Significant differences are marked \textsuperscript{$\dagger$} ($p < 0.05$).}
    \label{table-chat-statistics}
\end{table}

Table~\ref{table-chat-statistics} presents dialogue statistics, averaged over all topics for the base and enhanced LLM settings, respectively. See Appendix~\ref{apdx:dialogue-statistics} for topic-specific statistics. The results give insights into how engaged the participants were with the LLM, which is an indicator of explanation dialogue quality.

We observe that participants interacted significantly longer with the LLM in the enhanced setting, with an average duration of about 13 minutes compared to about 12 minutes in the base setting, and also send more queries (9.3 vs. 8.2). Furthermore, the LLMs' responses were, on average, shorter in the enhanced setting (12.0 vs. 17.8 sentences). The combination of longer interactions with more queries and shorter LLM responses suggests that the dialogues in the enhanced setting may have been more interactive and engaging. The fact that the participants' processing time remains similar across both settings further supports this assumption: While shorter answers of the enhanced LLM are likely to reduce reading time, the LLM may have encouraged the participants to spend more time thinking and formulating their next query.

\subsection{Monitoring}
\label{sec:co-constructive-behavior-results}

\begin{figure}[t]
    \centering
    \includegraphics[scale=1]{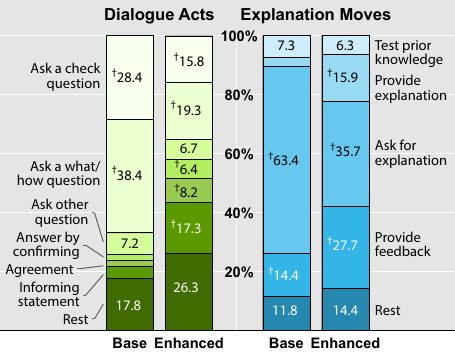}
    \caption{Proportions of annotated dialogue acts and explanation moves~\cite{wachsmuth2022wired, alshomary:2024} for the \emph{participants'} turns, normalized per dialogue of the \emph{base} and \emph{enhanced} settings, respectively. \emph{Rest} denotes the sum for all labels that have a proportion smaller than 5\% or that are too unspecific. Significant differences between the two settings are marked with \textsuperscript{$\dagger$} ($p < 0.05$). }
    \label{fig:turn-prediction-explainee}
\end{figure}

To further investigate the co-constructive behavior of the LLM and the resulting behavior of the participants, we automatically annotated the dialogue acts and explanation moves for the participants' and LLMs' turns in the dialogues.
Based on \citet{wachsmuth2022wired}, we focus on the following dialogue acts: (1) \emph{Ask a check question}, (2) \emph{Ask a what/how question}, (3) \emph{Ask other question}, (4) \emph{Answer by confirming}, (4) \emph{Agreement},
and (5) \emph{Informing statement}. For the explanation moves, we
consider: (1) \emph{Test prior knowledge},
(2) \emph{Provide explanation}, (3) \emph{Ask for explanation}, and (4) \emph{Provide feedback}. See Appendix~\ref{apdx:turn-level-prediction} for details.

Figure~\ref{fig:turn-prediction-explainee} shows the mean proportions of the annotated dialogue acts and explanation moves for the participants' turns normalized per dialogue of the two LLM settings, respectively. The proportions of the turns of the LLM are shown in Figure~\ref{fig:turn-prediction-explainer} in Appendix~\ref{apdx:turn-level-prediction}. The label \emph{Rest} is the sum of the proportions of the other labels introduced in \citet{wachsmuth2022wired} that have a proportion smaller than 5\% or that are too unspecific. 

We find that the LLMs in both settings primarily provide explanations and informing statements, as expected. However, in the enhanced setting, the LLM tests the participants' prior knowledge more frequently, and asks check and what/how questions more often. This behavior seems to influence the participants' behavior, as Figure~\ref{fig:turn-prediction-explainee} indicates: While the participants ask mostly check questions or what/how questions in both settings, their behavior is more diverse in the enhanced setting. The participants provide informing statements significantly more often, and they are more encouraged to also act as explainers and provide feedback.
This suggests that the enhanced setting can enable the LLM to apply more co-constructive behaviors, resulting in active participation of the participants.

\subsection{Scaffolding}
\label{sec:scaffolding-in-terms-of-Readability}
\begin{table}[t]
    \footnotesize
    \centering
    \renewcommand{\arraystretch}{0.95}
    \setlength{\tabcolsep}{3.5pt}
    
    \begin{tabular}{lrrr}
        \toprule
        \bf Topic & \multicolumn{1}{l}{\bf Gunning Fog $\downarrow$} & \multicolumn{1}{r}{\bf Type-Token $\downarrow$} & \multicolumn{1}{l}{\bf Entropy $\downarrow$} \\
        \midrule
        Quarto & +109.4\% {\tiny $\pm$ 100.9} & +23.0\% {\tiny $\pm$ \phantom{1}2.3} & +7.1\% {\tiny $\pm$ 8.9} \\
        Sleep & --10.1\% {\tiny $\pm$ \phantom{1}17.9} & +1.9\% {\tiny $\pm$ 12.6} & +0.8\% {\tiny $\pm$ 7.1} \\
        Black holes & --8.5\% {\tiny $\pm$ \phantom{1}16.7} & +6.0\% {\tiny $\pm$ 43.7} & --1.6\% {\tiny $\pm$ 5.3} \\
        \midrule
        Overall & +7.9\% {\tiny $\pm$ \phantom{1}58.5} & +7.3\% {\tiny $\pm$ 34.4} & +0.3\% {\tiny $\pm$ 7.1} \\
        \bottomrule
    \end{tabular}
    
    \caption{Change in average readability of LLM explanations (enhanced setting) before and after participants signaled non-understanding, measured in terms of  \textit{Gunning Fog} index, \textit{type-token} ratio, and Shannon \textit{entropy}. %
    }
    
    \label{table-complexity}
\end{table}

Next, we investigate the scaffolding behavior of the enhanced LLM, approximating it in terms of readability adjustment. In particular, we compare the readability of the LLM explanations before and after the participants signal non-understanding, based on the annotated explanation moves \emph{Provide explanation} and \emph{Signal non-understanding} \cite{wachsmuth2022wired}. We quantify readability in terms of text complexity (Gunning Fog Index), lexical diversity (Type-Token Ratio), and information density (Shannon Entropy). Ideally, all metrics should decrease after a participant signals non-understanding, given that we aim for the enhanced LLM to adapt its explanations based on the participant's current understanding.

Table~\ref{table-complexity} presents the percentage change in readability metrics for the LLM explanations before and after the participants signal non-understanding, averaged over 14 interactions (\textit{Quarto}: 2, \textit{Sleep}: 4, \textit{Black holes}: 8). We exclude the interactions with the LLM in the base setting, as only three dialogues included an annotated signal of non-understanding by the participants. We hypothesize that this difference does not reflect better understanding in the base setting, but rather that the enhanced LLM encourages the participants to more \emph{explicitly} signal their lack of understanding.

On average, the LLM in the enhanced setting successfully reduces its explanation complexity for two topics (--10.1\% for \emph{Sleep}, --8.5\% for \emph{Black holes}), while it unexpectedly increases complexity (+109.4\%) and lexical diversity (+23.0\%) for the topic \textit{Quarto}. The information density remains largely unchanged on average across all topics. The high standard deviation for all topics indicates that the scaffolding is not consistently successful. Possibly, the LLM did not accurately assess the current understanding. Alternatively, it may have accurately assessed the understanding but failed to provide appropriate scaffolding. Furthermore, the small sample size could explain the high standard deviation, particularly for the topic \textit{Quarto}. A qualitative analysis could provide further insights into these results, which we leave for future work.

\subsection{Explanatory Success}
\label{sec:explanatory-success}
\begin{table*}
    \small
    \centering
    \renewcommand{\arraystretch}{0.9}
    \setlength{\tabcolsep}{4.1pt}

    \begin{tabular}{llrrrrrrr}
        \toprule

        & & \multicolumn{2}{c}{\bf Motivation} & \multicolumn{2}{c}{\bf Subjective Compr.} & & & \\
        \cmidrule(l@{3pt}r@{3pt}){3-4}\cmidrule(l@{3pt}r@{3pt}){5-6}

        \bf Topic & \bf Setting & \multicolumn{1}{c}{\bf Pre}  & \multicolumn{1}{c}{\bf Post}  & \multicolumn{1}{c}{\bf Pre}  & \multicolumn{1}{c}{\bf Post} & \bf Objective Compr. & \bf Enabledness  & \bf Co-Construct. \\

        \midrule

        Quarto & Base & 3.9 {\tiny $\pm$ 0.5} & 4.1 {\tiny $\pm$ 0.6} & \textsuperscript{$\ddagger$}3.0 {\tiny $\pm$ 0.7} & \textsuperscript{$\dagger\ddagger$}3.8 {\tiny $\pm$ 0.6} & 73.3\% {\tiny $\pm$ 17.7} & 59.6\% {\tiny $\pm$ 24.8} & 3.8 {\tiny $\pm$ 0.6} \\
        & Enhanced & 4.0 {\tiny $\pm$ 0.6} & 4.1 {\tiny $\pm$ 0.6} & \textsuperscript{$\ddagger$}3.1 {\tiny $\pm$ 0.8} & \textsuperscript{$\dagger\ddagger$}4.0 {\tiny $\pm$ 0.6} & 73.9\% {\tiny $\pm$ 18.6} & 66.1\% {\tiny $\pm$ 18.1} & 4.0 {\tiny $\pm$ 0.7} \\
        [.3em]

        Sleep & Base & \textsuperscript{$\ddagger$}3.6 {\tiny $\pm$ 0.7} & \textsuperscript{$\ddagger$}4.0 {\tiny $\pm$ 0.7} & \textsuperscript{$\ddagger$}2.3 {\tiny $\pm$ 0.8} & \textsuperscript{$\ddagger$}3.9 {\tiny $\pm$ 0.5} & 75.7\% {\tiny $\pm$ 15.6} & 65.1\% {\tiny $\pm$ 22.8} & \textsuperscript{$\dagger$}3.8 {\tiny $\pm$ 0.6} \\
        & Enhanced & \textsuperscript{$\ddagger$}3.5 {\tiny $\pm$ 0.7} & \textsuperscript{$\ddagger$}4.0 {\tiny $\pm$ 0.5} & \textsuperscript{$\ddagger$}2.1 {\tiny $\pm$ 0.8} & \textsuperscript{$\ddagger$}3.8 {\tiny $\pm$ 0.5} & 74.5\% {\tiny $\pm$ 17.0} & 68.7\% {\tiny $\pm$ 21.9} & \textsuperscript{$\dagger$}4.1 {\tiny $\pm$ 0.5} \\
        [.3em]

        Black holes & Base & 3.6 {\tiny $\pm$ 0.7} & 3.8 {\tiny $\pm$ 0.8} & \textsuperscript{$\ddagger$}2.0 {\tiny $\pm$ 0.8} & \textsuperscript{$\ddagger$}3.5 {\tiny $\pm$ 0.6} & 76.2\% {\tiny $\pm$ 12.9} & 76.4\% {\tiny $\pm$ 20.3} & \textsuperscript{$\dagger$}3.6 {\tiny $\pm$ 0.7} \\
        & Enhanced & \textsuperscript{$\ddagger$}3.7 {\tiny $\pm$ 0.7} & \textsuperscript{$\ddagger$}3.9 {\tiny $\pm$ 0.7} & \textsuperscript{$\ddagger$}2.2 {\tiny $\pm$ 0.9} & \textsuperscript{$\ddagger$}3.7 {\tiny $\pm$ 0.6} & 75.2\% {\tiny $\pm$ 14.0} & 73.6\% {\tiny $\pm$ 24.4} & \textsuperscript{$\dagger$}4.2 {\tiny $\pm$ 0.5} \\

        \midrule

        Overall & Base & \textsuperscript{$\ddagger$}3.7 {\tiny $\pm$ 0.7} & \textsuperscript{$\ddagger$}3.9 {\tiny $\pm$ 0.7} & \textsuperscript{$\ddagger$}2.4 {\tiny $\pm$ 0.9} & \textsuperscript{$\ddagger$}3.7 {\tiny $\pm$ 0.6} & 75.1\% {\tiny $\pm$ 15.6} & 67.0\% {\tiny $\pm$ 23.8} & \textsuperscript{$\dagger$}3.7 {\tiny $\pm$ 0.6} \\
        & Enhanced & \textsuperscript{$\ddagger$}3.7 {\tiny $\pm$ 0.7} & \textsuperscript{$\ddagger$}4.0 {\tiny $\pm$ 0.6} & \textsuperscript{$\ddagger$}2.5 {\tiny $\pm$ 1.0} & \textsuperscript{$\ddagger$}3.8 {\tiny $\pm$ 0.6} & 74.6\% {\tiny $\pm$ 16.6} & 69.5\% {\tiny $\pm$ 21.9} & \textsuperscript{$\dagger$}4.1 {\tiny $\pm$ 0.6} \\

        \bottomrule
    \end{tabular}

    \caption{Results of the questionnaires in terms of \textit{motivation}, \textit{subjective comprehension} (both before and after the interaction), \textit{objective comprehension}, \textit{enabledness}, and \textit{co-constructive} behavior of the LLM. All numbers are averaged per topic over all participants in each setting. Significant differences between settings are marked with \textsuperscript{$\dagger$} ($p < 0.05$), significant differences between pre- and post-interaction metrics are marked with \textsuperscript{$\ddagger$} ($p < 0.05$).}

    \label{table-performance}
\end{table*}

To examine whether the LLMs' co-constructive behavior affects the success of the explanations, we analyze the results of the understanding questionnaires.
In Table~\ref{table-performance}, for both LLM settings, the \emph{motivation}, \emph{subjective comprehension}, and \emph{co-constructiveness} are shown as the average value over all statements of the respective questionnaire, ranging from $1$ (lowest) to $5$ (highest). The values for \emph{objective comprehension} (14 statements) and \emph{enabledness} (five questions) represent the average percentage of correctly answered statements/questions.

Across all topics and settings, participants reported, on average, an increased motivation to engage with the explanandum after the interaction, e.g., from 3.5 to 4.0 in the case of \textit{Sleep} in the enhanced setting. Subjective comprehension also increased significantly, from approximately 2.0--3.1 before the interaction to 3.5--4.0 afterward, regardless of the setting. These results suggest that in both settings, interacting with the LLMs increases the participants' motivation and confidence in their understanding of the explanandum in a similar way.

When comparing the objective understanding, on average, the participants’ objective comprehension and enabledness after the interaction were also found to be similar in both settings. This indicates that the enhanced setting does not seem to lead to significantly better explanations or greater understanding than the base setting.

However, the participants rated the enhanced LLM as more co-constructive (4.1 vs. 3.7), aligning with the findings in Section~\ref{sec:co-constructive-behavior-results}.
Interestingly, this increased co-constructiveness, on average, did not translate into a higher objective understanding in the enhanced setting than in the base setting. This contradicts our expectations that co-constructive behavior improves explanatory success. Due to the consistent findings in previous studies \citep{forbes-riley2011,robrecht2023}, we hypothesize that this is a shortcoming of the LLM's capabilities (as discussed in Section~\ref{sec:scaffolding-in-terms-of-Readability}), rather than evidence against the effectiveness of co-constructive explanations. Furthermore, we find a significant positive correlation between the prior motivation and assessed co-constructive behavior (Kendall's $\tau = 0.19$) that suggests that the participants need to be motivated to experience good co-constructive behavior of LLMs.

\subsection{Objective Understanding}
\label{sec:objective-comprehension-spread}

\begin{figure}
    \centering
    \includegraphics[width=0.48\textwidth]{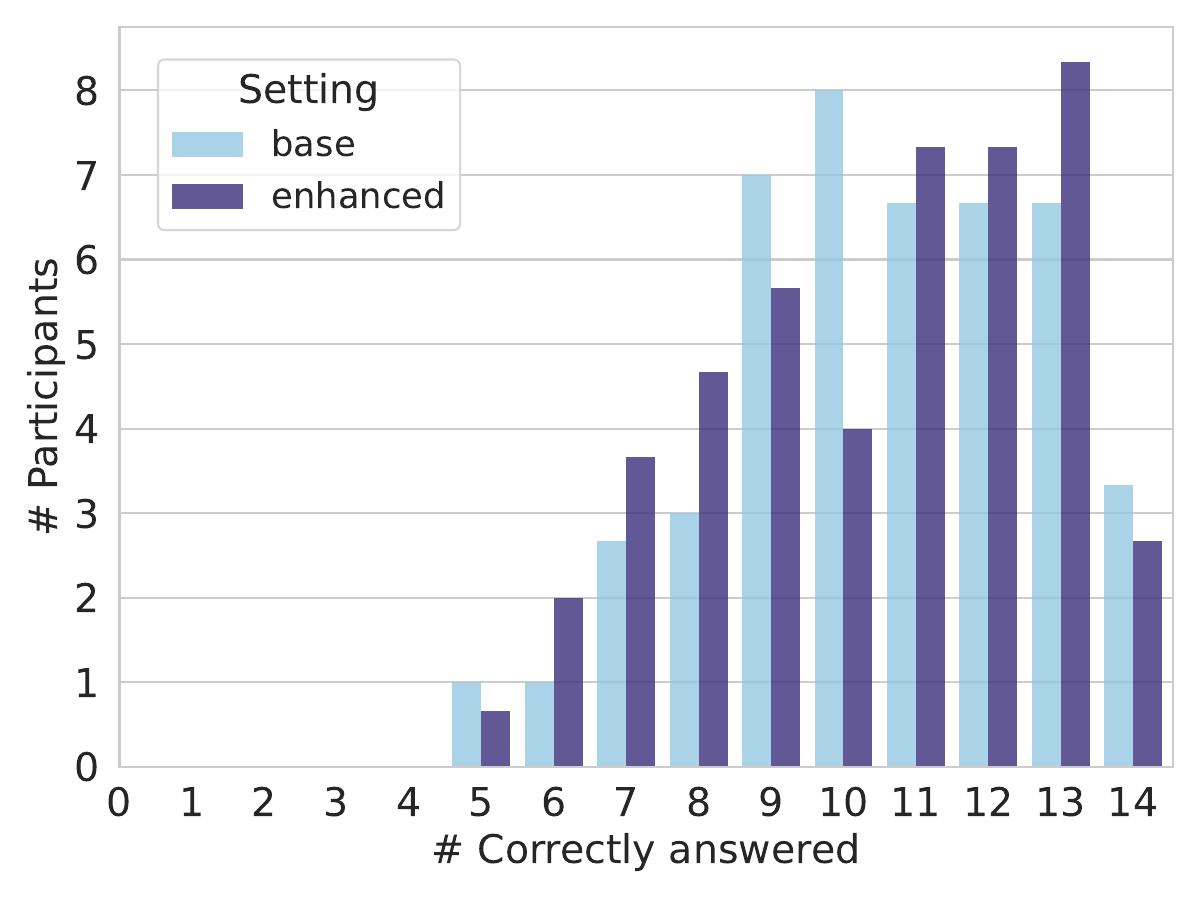}
    \caption{Results of the post objective comprehension questionnaire (14 statements) averaged over all topics for the \emph{base} and \emph{enhanced} settings, respectively.}
    \label{fig:obj-comprehension-histogram}
\end{figure}

To better understand the explanatory success, we investigate the participants' objective understanding in detail, evaluating how many questions were answered correctly per participant.

Figure~\ref{fig:obj-comprehension-histogram} shows the results of the objective comprehension questionnaire, averaged across all topics. While the scores in the base setting follow approximately a shifted normal distribution, the enhanced setting shows increased tails on both ends of the histogram: Fewer participants had an average of 10 correct answers, while more participants performed better or worse. This finding suggests that the co-constructive behavior of the enhanced LLM can have a positive effect on the participants’ objective comprehension. However, this effect is not consistent across all participants and is therefore hidden in the average values in Table~\ref{table-performance}. The participants' performance on the enabledness questionnaire shows a comparable trend.

To further explore reasons for this distribution spread, we determine whether the objective comprehension correlates with certain dialogue acts and explanation moves for the enhanced setting.

For the LLM turns, there is a significant positive correlation between objective comprehension and occurrences of the dialogue act \emph{Provide informing statement} ($\tau = 0.15$) as well as the explanation move \emph{Provide explanation} ($\tau = 0.17$). While this suggests that more explanations from the LLM may lead to a better objective comprehension, it also shows that other co-constructive behaviors may not have led to an increased comprehension.

For the participant turns, there is a similar significant positive correlation between objective comprehension and \emph{Provide explanation} ($\tau = 0.16$). This might indicate that the more the LLM encourages participants to provide explanations themselves, the better their objective comprehension of the explanandum becomes. This is in line with research showing that self-explanations can enhance understanding~\cite{chi1994,fiorella2023}. The fact that the enhanced LLM does not consistently engage the participants to provide explanations could explain the slight skew in Figure~\ref{fig:obj-comprehension-histogram}.

Overall, the results indicate that the evaluated LLM shows certain capabilities to lead co-constructive explanation dialogues as explainers, which can lead to improved understanding. This is consistent with the findings that more co-constructive dialogues increase explanatory success~\cite{robrecht2023}. However, while the LLM can simulate monitoring to some extent, scaffolding success seems to be rather inconsistent.

\section{Qualitative Analysis}
\label{sec:quali-analysis}
In addition to our quantitative analysis, we examine nine dialogues qualitatively to gain deeper insight into the LLM's co-constructive behavior. Appendix~\ref{apdx:qualitative-analysis} shows the dialogue selection and excerpts referenced here by participant ID (PID).

Co-constructive explanations rely on the explainee's active participation. In the base setting, we find that the interactions often follow a teacher-student setting, with the LLM providing long monologues and minimal opportunities for active participation (e.g., PID 4be6). This challenged the processing capacity of some participants (e.g., PID 4be6, \emph{Yeah, this is a lot can we take it bit by bit}). The enhanced setting yields evidence of better monitoring by assessing the explainee's understanding and assigning tasks to reproduce information that has been explained previously (e.g., PID 43b6). In both settings, however, the explainee often has to invest extra effort to enable monitoring and scaffolding from the LLM by answering multiple preference questions (e.g., PID 943a, \emph{Would you like to know more about [...]?}), leading to a misalignment with the explainee's needs (e.g., PID 943a, 1570). Our asynchronous unimodal chat-like setup likely contributes to this issue by limiting backchanneling of implicit and unconscious signals, that are typically present in synchronous multimodal interactions~\cite{inden2013backchanneling, goodwin2018multimodal}.

We further observe behavioral differences across topics. The topic \textit{Black holes} led to more monologues that the explainees consumed passively, possibly due to its factual nature. In contrast, \textit{Sleep} and \textit{Quarto} seemingly also allow the explainee to talk about personal experiences (e.g., PID 2417) or made-up game situations (e.g., PID 943a).

Overall, we observe surface-level adaptation capabilities in the enhanced setting. However, the adaptation appears static due to seemingly incomplete scaffolding capabilities of the LLM. In line with our quantitative analysis, the findings emphasize that the out-of-the-box capabilities of the LLM benefit motivated participants, but not everyone.

\section{Conclusion}
\label{sec:conclusion}

In this paper, we have investigated to what extent out-of-the-box LLMs behave co-constructively in explanation dialogues as well as how this behavior may improve the success of generated explanations. Through a user study in which human explainees interacted with an LLM explainer, we have evaluated how well LLMs can follow instructions to apply co-constructive behaviors (monitoring and scaffolding) while explaining a topic to an explainee.

In our quantitative and qualitative analyses of the dialogues and the participants' understanding of the topic, we find that explicit prompting can enable the evaluated LLM to explain co-constructively \emph{to some extent}. On the one hand, it actively utilized co-constructive patterns common to monitoring and scaffolding behavior, such as asking verification questions, adjusting the readability of explanations, and encouraging active participation. These patterns show potential to increase the explainees' understanding. On the other hand, the patterns are used inconsistently and rather statically, instead of being adjusted to the explainee's needs.

Overall, the evaluated LLM shows promising results that can serve as a good basis to successfully lead explanation dialogues, but further advancements are needed to enable real co-construction. In particular, we expect that truly co-constructive LLMs not only engage users in a dialogue but also dynamically adapt through consistently successful monitoring and scaffolding to ensure explanatory success. The insights of this work contribute to the understanding of the adaptivity of LLMs and thus define a starting point for further work on co-constructive approaches in XAI.

\section*{Limitations}
\label{sec:limitations}

As this study is an early evaluation on aspects of co-constructive behavior of LLMs, some limitations should be considered when interpreting the results.

First, we focus on a single LLM, which may limit the generalizability of our results. We do think, however, that the LLM represents a common choice at the time of conducting the experiments, leading us to believe that similar results may be obtainable with other open-weight models.

Second, our unimodal setup limits the ability to convey and interpret signals with other modalities, such as intonation and facial expressions. For example, it may be harder to assess the explainee's emotional state, as discussed in Section~\ref{sec:quali-analysis}. Thereby, the setup generally limits the co-constructive behavior that can emerge.

Third, the asynchronous chat-like setup imposes certain limitations on the interaction between the LLM and the participant. For example, it does not allow for interruptions, and makes ``chit-chat'' inconvenient; again, due to the bigger effort of typing compared to speaking, which may, in some cases, be important for social aspects of interaction.

Despite the limitations, we nevertheless think that the choice of a unimodal chat-like setup was justified. First, since this is an early study on LLMs in this direction, a unimodal setting reduces the number of variables that might influence the results and complexity of the analysis. Furthermore, at the time of conducting the study, LLMs have been commonly used in unimodal settings. Our evaluation is thus representative of the current de facto standard for users to interact directly with LLMs, e.g., in chatbot applications such as ChatGPT.

Lastly, we point to the inability of our study to control for the participants' intrinsic motivation for the topic, which may have led to a variability in engagement and learning outcomes, independent of the LLMs' co-constructive behaviors. However, we tried to control for extrinsic motivation by paying participants a bonus if they showed actual interest in learning more about the provided topic throughout the dialogue (see the next section for payment details). We semi-automatically evaluated each participant's chat and paid the bonuses accordingly.

\section*{Ethical Considerations}
\label{sec:ethics}

Co-constructiveness, the subject of our experiments, promises to provide notable benefits to users through more personalized and effective explanations. If working effectively, co-constructive LLMs may enable explanations adapted to one's needs and state of understanding. Subjects of such explanation cannot only be topics like as \textit{Quarto} and \textit{sleep}, but also predictions and decisions of AI models. A co-constructive LLM may, therefore, positively impact the explainability of AI models.

However, since LLMs are not perfectly accurate at all times, such explanations can also provide wrong or incomplete information. This opens the potential for the user to trust in false information generated by the LLM, if it is not verified further. However, our approach is applicable to any instruction-tuned LLM. In the future, it could thus easily be adapted to LLMs that have been optimized to not generate false information.

Lastly, we acknowledge that we only consider Standard American English (SAE) in this study. As previous studies have shown, current state-of-the-art LLMs tend to work better for SAE than for dialects \cite{ziems2022,ondrejova2024}. This has the potential to negatively impact dialect speakers. As our focus was to create an understanding of the potential co-constructive behavior of current LLMs, evaluating for more than a single language was outside our scope. We do, however, encourage future research to facilitate fairness and equality in generated explanations.

We conducted our study on \textit{Prolific}. We estimated the study to take a maximum of 30 minutes per participant and paid at least \pounds6, thus targeting a rate of \pounds12/hour. In order to obtain high-quality dialogues, an additional bonus of \pounds1.50 was available for those who demonstrated a high level of engagement and motivation to learn about the topic during the interaction. In addition to ensuring data quality, we used two attention checks (instructional manipulation checks) in the first questionnaire following Prolific's policy. We clearly communicated those details in the study description. Before participation, participants were presented with an informed consent form that clearly stated that anonymized excerpts of their dialogues and responses may be used to illustrate findings in research publications.

\section*{Acknowledgments}

This work has been supported by the ``HybrInt - Hybrid Intelligence through Interpretable AI in Machine Perception and Interaction'' project (Zukunft Nds, Niedersächsisches Ministerium für Wissenschaft, Grant ID: ZN4219), the Deutsche Forschungsgemeinschaft (DFG, German Research Foundation) under project number TRR 318/1 2021 – 438445824, and the Federal Ministry of Education and Research (BMBF), Germany under the AI service center KISSKI (grant no. 01IS22093C).

We thank Thilo Glißmann, physicist at the Fraunhofer Institute in Kassel, and Dr. med. Katrin Meyer, head of the sleep laboratory at the Medizinische Hochschule Hannover (MHH), for validating our objective understanding questionnaires. 
The programming was supported by ChatGPT.

\bibliography{co-constructive-llms}

\newpage
\appendix
\section{Experimental Details}
\label{apdx:experimental-details}

\subsection{Complete system prompts}
\label{apdx:full-prompts}

Below, we provide the full system prompts for both, the base setting and the enhanced setting, used to instruct the LLM. While we did not conduct a comprehensive study to optimize the prompts, particular details, such as emphasizing important aspects with capitla letters, were determined in small pilot experiments.

\paragraph{Base setting}

For the baseline setting, we use a minimal prompt, only including a task description that clarifies the context, as described in Section~\ref{sec:cocon-llms}. More specifically, we use the following prompt:

\begin{flushleft}
\small
\setstretch{1.05}
{\ttfamily
    You act as the explainer in a chat environment to explain a specific topic to the user chosen by the user.
}
\end{flushleft}

\paragraph{Enhanced setting}

As described in Section~\ref{sec:cocon-llms}, for the enhanced setting, we use a more detailed prompt that, in addition to the task and context description, instructs the LLM to apply co-constructive behavior. More specifically, we instruct the LLM to make use of monitoring and scaffolding, and also include a definition of both. The following is the full prompt for this setup:
\begin{flushleft}
\small
\setstretch{1.05}
{\ttfamily
    You act as the explainer in a co-constructive explanation chat environment to explain a specific topic to the user chosen by the user.\linebreak
    
    You apply monitoring and scaffolding techniques to enable the user in the topic the user asks you to explain.

    You DO NOT make it explicit that you apply monitoring and scaffolding.
    You DO NOT suggest any topics.\linebreak

    Definition of monitoring: Through monitoring, the explainer aims to identify the knowledge gap through diagnostic queries (a recurring task throughout the dialogue) and verification questions in a dialogue. Monitoring allows the explainer to evaluate whether the explainer's way of explaining has been successful or whether further elaboration or modification of the explanation is needed. \linebreak

    Definition of scaffolding: Scaffolding describes the process and actions of the explainer to adjust the dialogue and explanations, based on the information gathered during the monitoring; both, monitoring and scaffolding, happen in accordance with each other. Scaffolding actions can, for example, be to keep the explanans digestible and adjust their complexity, or providing further context for explanations, based on dialogue history and the outcome of the verification processes performed during the monitoring.
}
\end{flushleft}

\subsection{Language Model Hyperparameters}
\label{apdx:hyperparamters}

As detailed in Section~\ref{sec:chat-application}, we use the instruction-tuned variant of the open-weight model Llama 3.1 with 70 billion parameters \cite{dubey2024} as provided via the Chat AI API platform of the GWDG \cite{doosthosseini2024}. We apply a temperature of $1.0$, and set top-p value to $1.0$.

\subsection{User study}
\label{apdx:study-application}

\begin{figure}
    \centering
    \includegraphics[width=\linewidth]{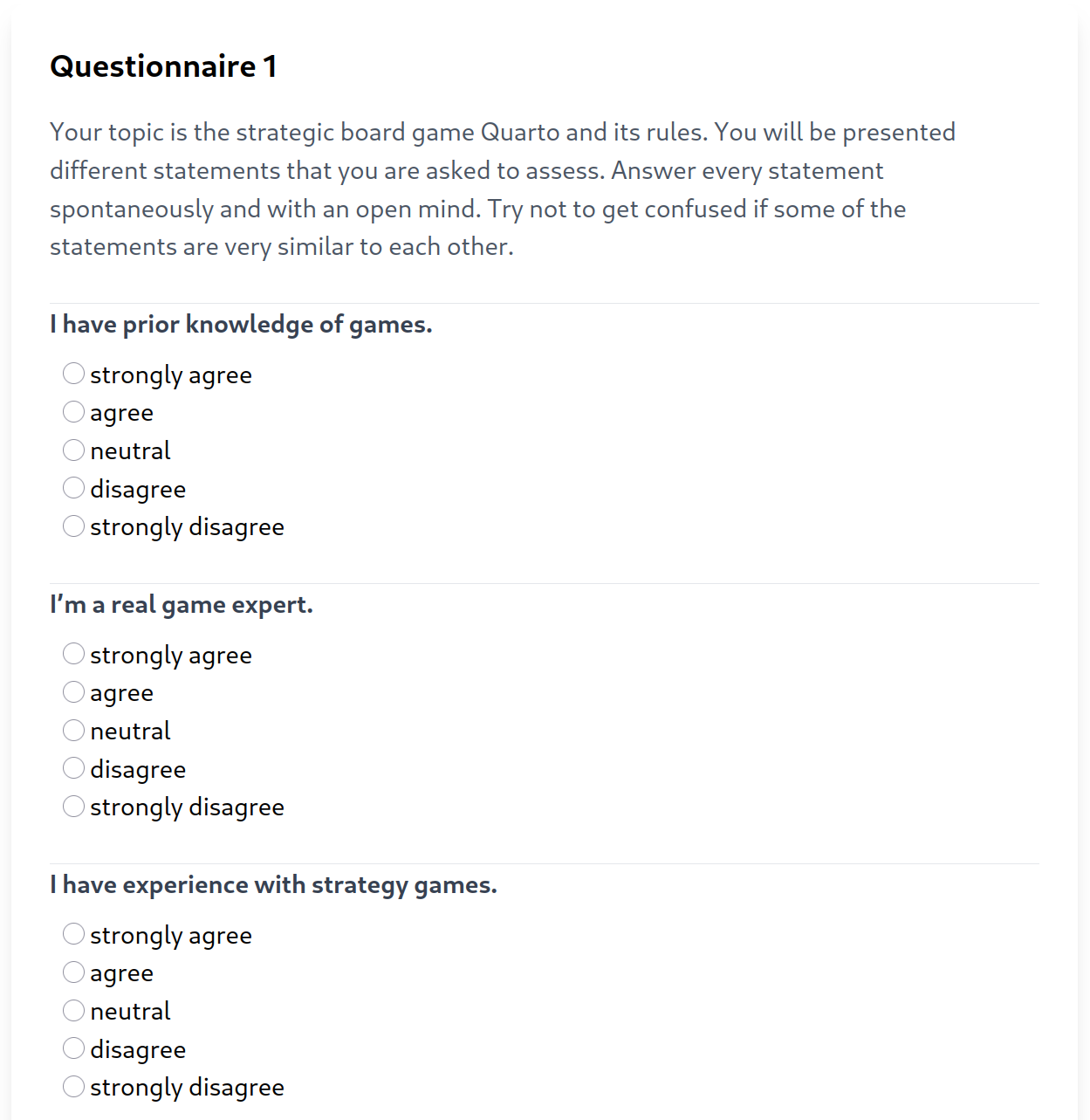}

    \caption{The study application interface for answering questionnaires, here exemplified for the first questions of the subjective comprehension questionnaire.}
    \label{fig:study-application-interface-questionnaires}
\end{figure}

\begin{figure}
    \centering
    \includegraphics[width=\linewidth]{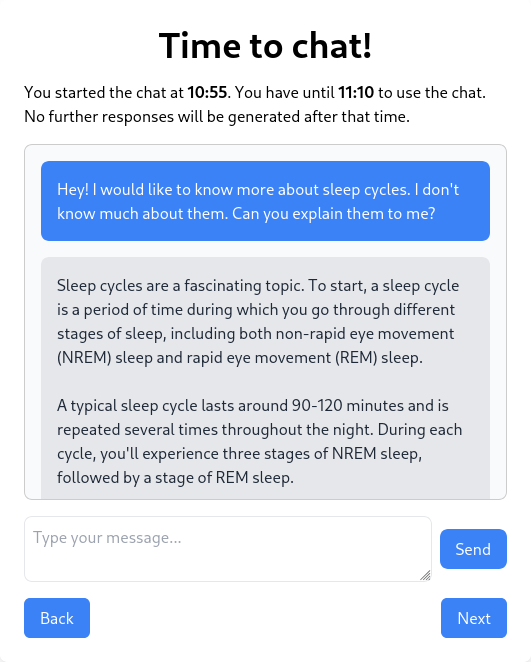}

    \caption{The study application interface for interacting with the LLM. After the first user message is sent, a timer of 15 minutes is started that limits the maximum interaction time with the LLM. See Section~\ref{sec:user-study} for details.}
    \label{fig:study-application-interface-chat}
\end{figure}

To conduct our user study, we developed an application based on the Django framework
\footnote{The application can be found under \url{https://github.com/webis-de/sigdial25-co-constructive-llms}}, which guides the user through multiple questionnaires and an LLM interaction screen. Figure~\ref{fig:study-application-interface-questionnaires} and Figure~\ref{fig:study-application-interface-chat} show screenshots of the questionnaire interface and the LLM interaction interface, respectively.

Before each participant started the study, they were informed that the study would involve a chat-based interaction to explore a particular topic, followed by five questionnaires before and after the interaction. They were told that the study would take approximately 30 minutes to complete. By clicking the "Agree and Start" button, participants provided consent and acknowledged that anonymized chat excerpts and responses could be used in research publications. In addition, it was emphasized that all data would remain confidential and would only be used for research purposes. Participants were also informed that the study includes attention checks.

Before starting the chat with the LLM, the participants were informed that their task now is to chat about the explanandum given to them beforehand. They were notified that they could chat for a maximum of 15 minutes. To maintain engagement and ensure high-quality interactions, we informed them that two post-interaction questionnaires will evaluate their understanding of the explanandum. In addition to controlling for extrinsic motivation, we told them they would get a \pounds1.50 bonus for actively engaging in the chat and showing real motivation to learn. To avoid biasing the participants, we did not reveal that the study focuses on co-constructive explanation dialogues.%
\footnote{The data of our user study can be found under \url{https://github.com/webis-de/sigdial25-co-constructive-llms-data}}

\subsection{Selection of Initial Explananda}
\label{apdx:global-explanada-selection}

To minimize the potential influence of external factors on the explanation dialogue, we select the initial explananda based on five main criteria: (1)~The topic should not require a lot of background knowledge so that it can be grasped in the limited chat time. However, the topic should be complex enough to prevent complete understanding by all participants, ensuring valuable insights from the explanation processes; (2)~The topic should be unfamiliar to most people, allowing us to recruit enough non-expert participants, but common enough so that an ``out-of-the-box'' LLM is be able to generate feasible explanations; (3)~The topic should be universally relevant to all people, regardless of demographic factors, such as gender, age, or ethnicity to avoid exclusion of minorities (as far this is possible to assess); (4)~The initial explanandum should be well defined so that participants do not drift off into unrelated areas; (5)~The explanandum should allow for an evaluation of the participants' comprehension and enabledness to allow us to assess the understanding gained through the interaction.

\subsection{Questionnaires}
\label{apdx:questionniare-details}

As detailed in Section~\ref{sec:questionnaires}, we use a combination of pre-interaction and post-interaction questionnaires to evaluate several aspects.%
\footnote{The questionnaires used in our user study can be found under \url{https://github.com/webis-de/sigdial25-co-constructive-llms}}

\begin{figure}
    \centering

    \includegraphics[width=0.459\textwidth]{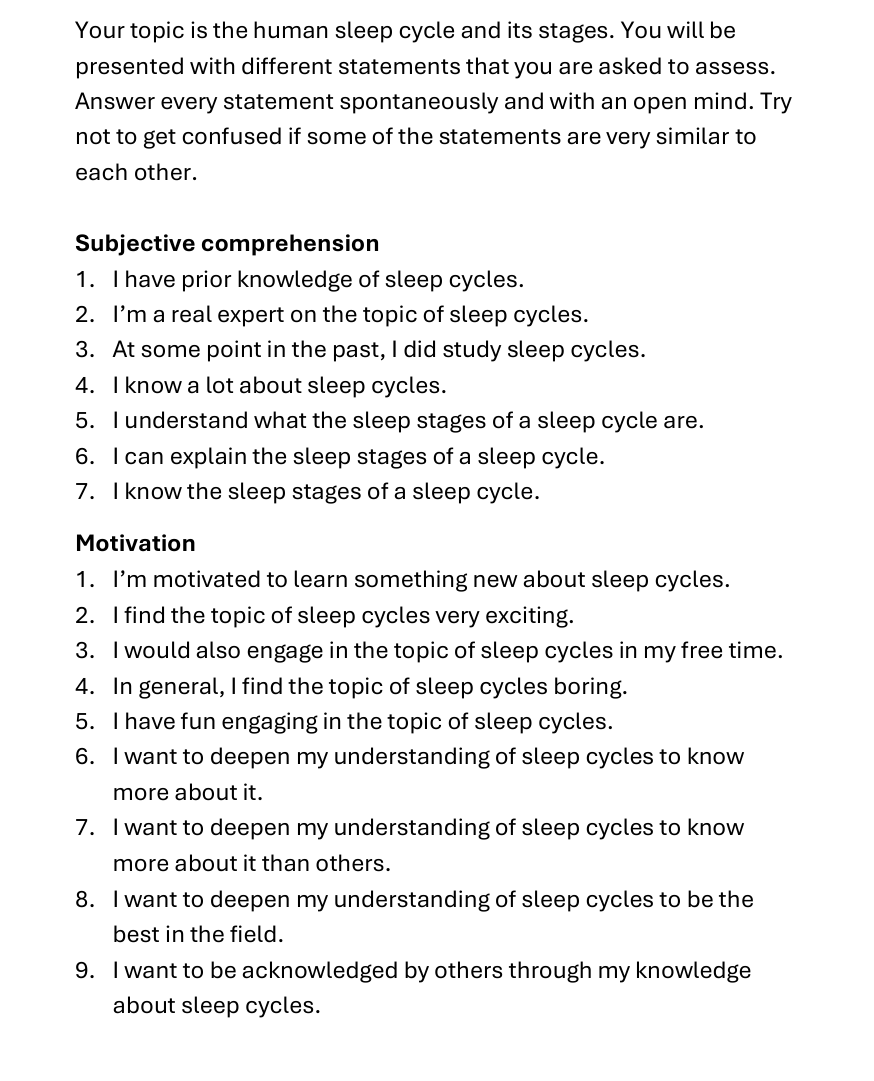}

    \caption{The questionnaire to assess the participants' prior subjective comprehension of \textit{sleep cylces} and the motivation to learn about it before interacting with the LLM. The statements are based on \citet{buhl2025partnermodelquestionnaire} and are rated on a five-point Likert scale.}
    \label{fig:sleep-subjective-comprehension-questionnaire-pre}
\end{figure}

\begin{figure}
    \centering

    \includegraphics[width=0.439\textwidth]{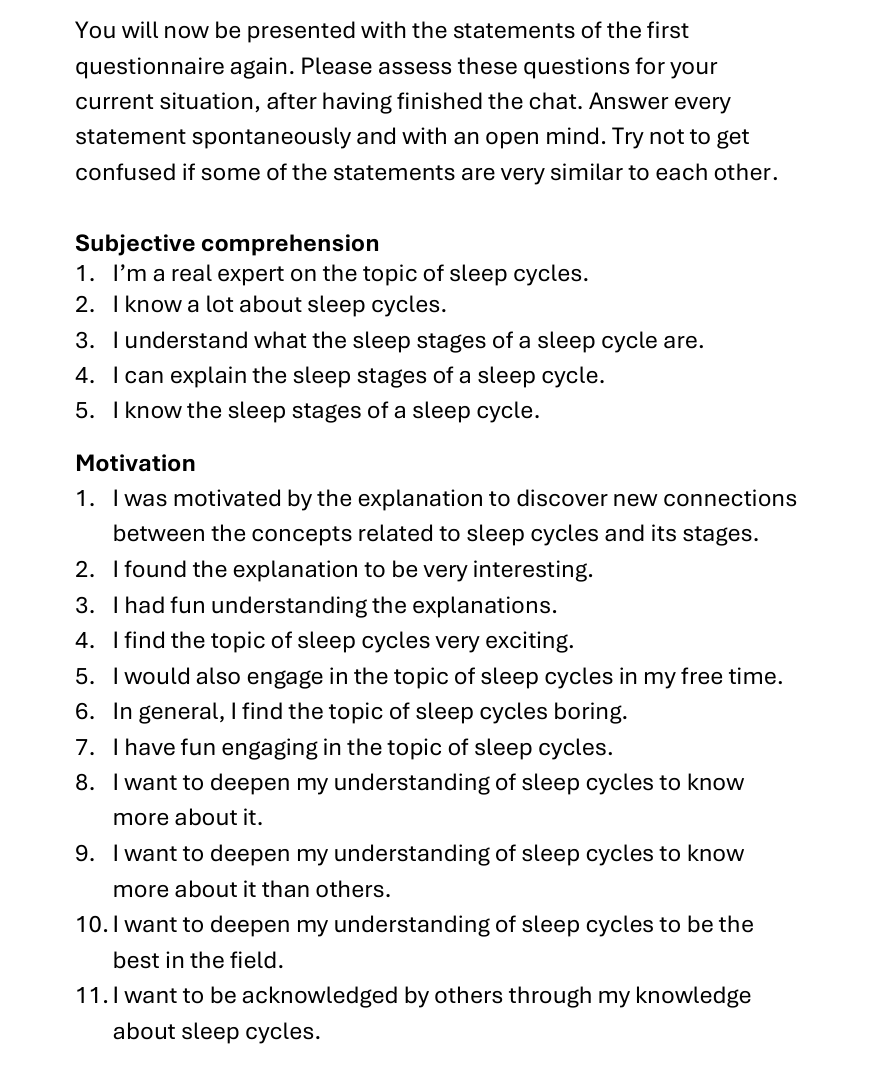}

    \caption{The questionnaire to assess the participants' post subjective comprehension of the explanandum related to \textit{sleep} and the motivation to learn about it after interacting with the LLM. The statements are based on \citet{buhl2025partnermodelquestionnaire}. The participants rate the statements on a five-point Likert scale.}
    \label{fig:sleep-subjective-comprehension-questionnaire-post}
\end{figure}

Before the interaction with the LLM, the participants are asked to complete a subjective comprehension questionnaire with 16 statements, each rated on a five-point Likert scale ranging from ``strongly agree'' to ``strongly disagree''. The statements ask about the participant's self-assessed comprehension of the explanandum, interest in the explanation, and the topic, as well as the extrinsic motivation for the topic (see Figure~\ref{fig:sleep-subjective-comprehension-questionnaire-pre} for an example on the topic \textit{sleep}). Participants were asked to complete the same questionnaires after the interaction with the LLM to assess pre- and post-interaction comprehension. Some statements present in the pre-interaction questionnaire were, however, not present in the post-interaction questionnaire, as it does not make sense to ask for general statements twice. Furthermore, some statements are rephrased to match the tense (see Figure~\ref{fig:sleep-subjective-comprehension-questionnaire-post} for an example). Both questionnaires were derived from \citet{buhl2025partnermodelquestionnaire} and adapted to our explananda.

\begin{figure}
    \centering

    \includegraphics[width=0.439\textwidth]{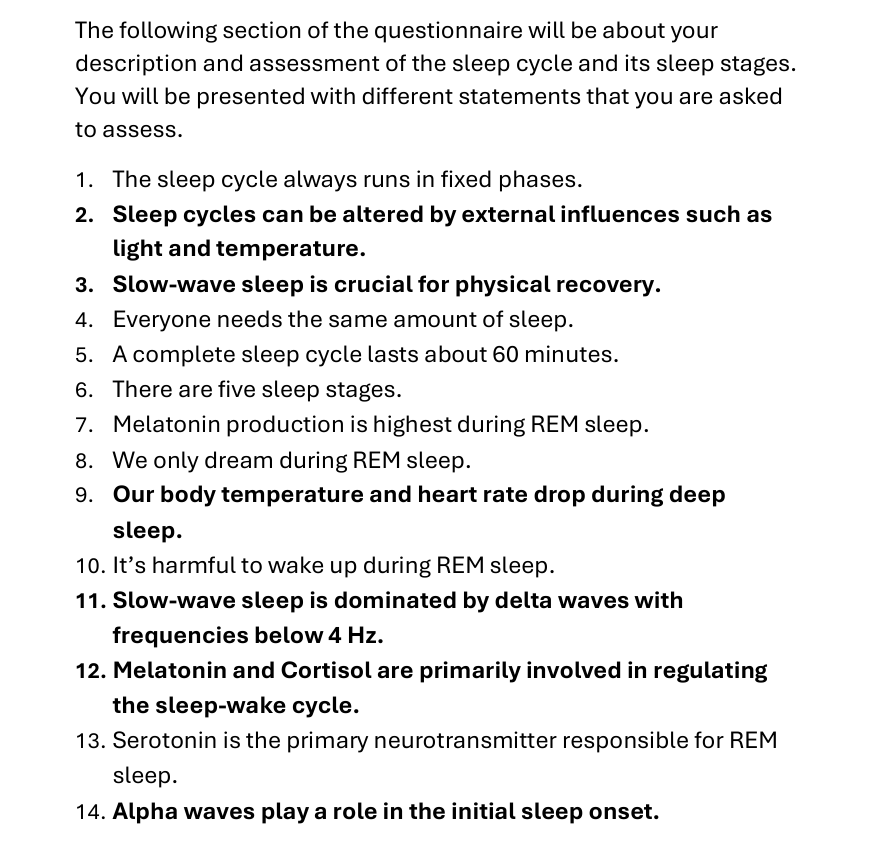}

    \caption{The questionnaire to assess the participants' objective comprehension of the explanandum related to \textit{sleep} after interacting with the LLM. The statements are based on \citet{terfloth2025}. The participants validate the statements to be either \textit{correct} or \textit{not correct}. The correct statements are marked in bold.}
    \label{fig:sleep-obejctive-comprehension-questionnaire}
\end{figure}

\begin{figure}
    \centering

    \includegraphics[width=0.439\textwidth]{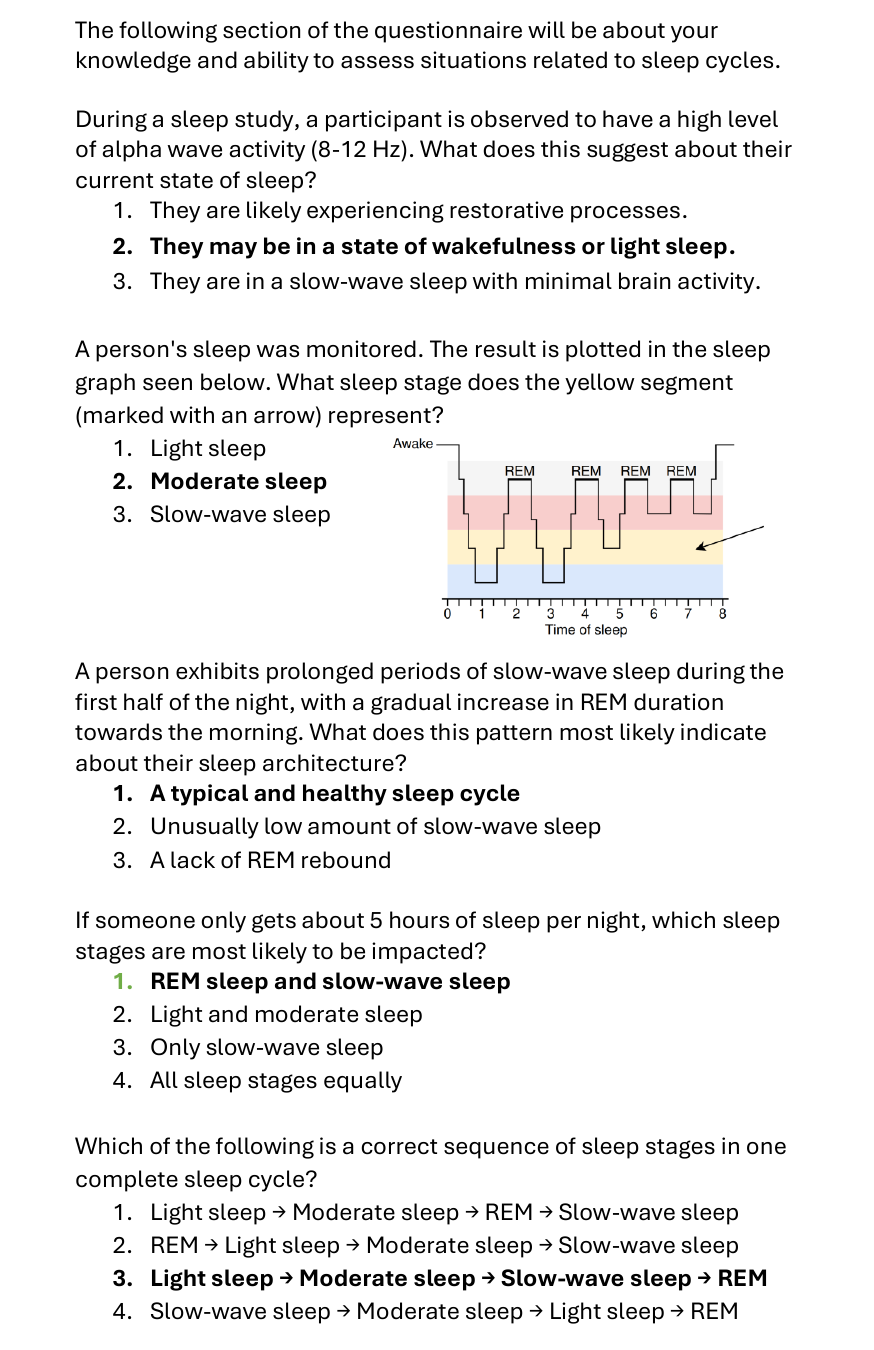}

    \caption{The questionnaire to assess the participants' enabledness of the explanandum related to \textit{sleep} after interacting with the LLM. The questions are based on \citet{terfloth2025}. For every question, the participants chose the correct answer. The correct answers are marked in bold for overview purposes only.}
    \label{fig:sleep-enabledness-questionnaire}
\end{figure}

After the interaction with the LLM, participants completed two additional questionnaires regarding their understanding: an objective comprehension questionnaire (see Figure~\ref{fig:sleep-obejctive-comprehension-questionnaire} for an example on the topic \textit{sleep}) and an enabledness questionnaire (see Figure~\ref{fig:sleep-enabledness-questionnaire} for an example on the topic \textit{sleep}). The objective comprehension questionnaire consists of 14 statements that the participant can either agree or disagree with, and is supposed to assess their actual comprehension of the explanandum. The enabledness questionnaire contains five multiple-choice questions that evaluate their ability to apply their knowledge to practical situations related to the explanandum. Both questionnaires are derived from \citet{terfloth2025} and adapted to our explananda. We only test for objective understanding after the interaction with the LLM to prevent biasing the participants. To draw conclusions about the understanding gained during the interaction, two different questionnaires of equal complexity would be required. As we based our study on existing questionnaires, this is out of scope for our study.

At the end of each post-interaction understanding questionnaire, we added an open question to ask if the explanations of the LLM were sufficient to answer the statements/questions. The answer to this open question should allow us to evaluate whether the participant used external sources to complete the two questionnaires. However, answering the open questions was not mandatory.

\begin{figure}
    \centering

    \includegraphics[width=0.439\textwidth]{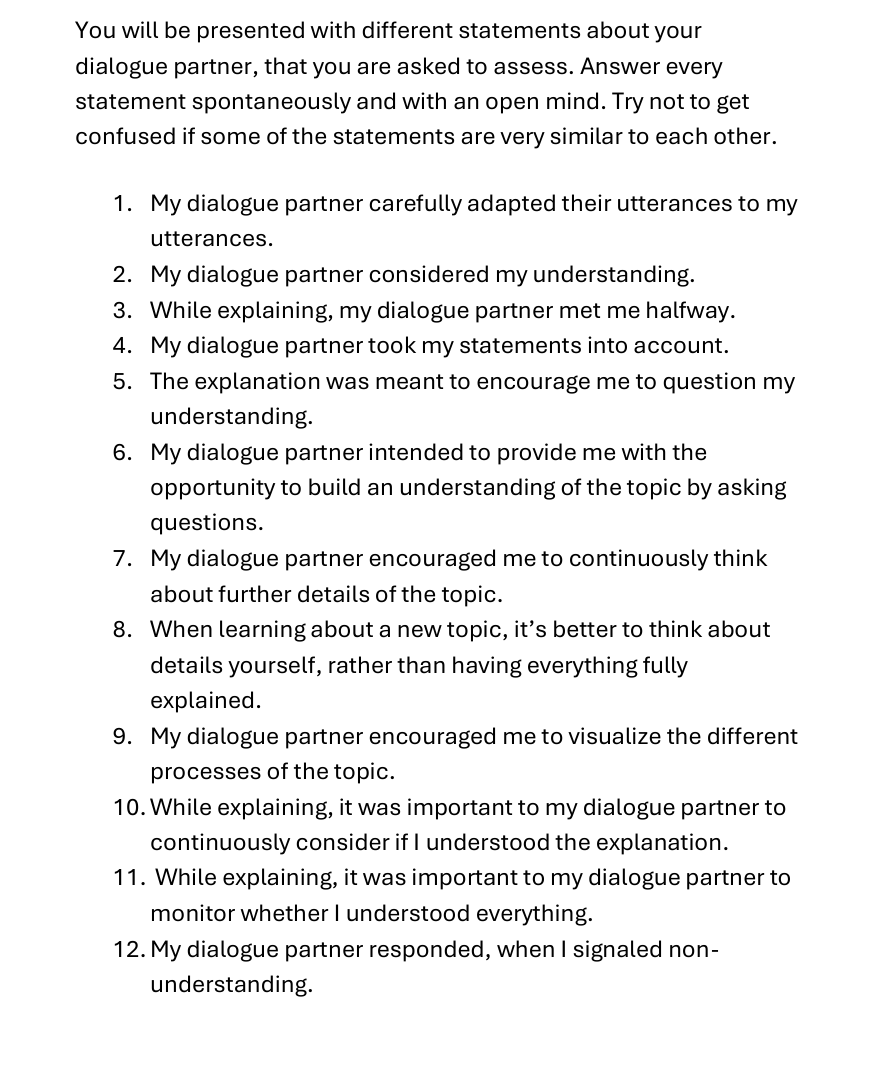}

    \caption{The questionnaire to assess the co-constructive behavior of the LLM. The items are adopted from \citet{buhl2025}. The participants rate the statements on a five-point Likert scale.}
    \label{fig:co-construction-questionnaire}
\end{figure}

Finally, the participants completed a last questionnaire to measure the co-constructive behavior \citet{buhl2025} of the LLM during the interaction. Figure~\ref{fig:co-construction-questionnaire} shows an overview of all items of our questionnaire. Since the eighth statement is not directly related to co-constructive behavior, we exclude this statement when calculating the co-constructiveness average for Table~\ref{table-performance}.

\subsection{Dialogue Act and Explanation Move Prediction Setup}
\label{apdx:turn-level-prediction-setup}

\citet{wachsmuth2022wired} and \citet{alshomary:2024} annotate the turns of a dialogue for the \textit{ELI-5} \cite{fan2019} corpus and transcripts of the freely available \textit{5-levels} video series published by Wired%
\footnote{\url{https://www.wired.com/video/series/5-levels}, accessed on 2025-04-11.}.
They annotate three aspects: the explanation move, the dialogue act, and the topic of each turn. The best overall performance of a classifier to automatically predict those three aspects was achieved by training on both corpora together \cite{alshomary:2024}. We follow this approach using the code published by \citet{alshomary:2024} to train a classifier on both corpora. In contrast to~\citet{alshomary:2024}, we use the pre-trained \textit{Longformer} encoder model \cite{beltagy2020}, since our turns can be notably longer.

\begin{table}
    \centering
    \renewcommand{\arraystretch}{0.9}
    \small
    
    \begin{tabular}{llrr}
        \toprule
        \textbf{Topic} & \textbf{Setting} & \textbf{Before} & \textbf{After} \\
        
        \midrule

        Quarto 		& Base     & 49 & 46 \\
         			& Enhanced & 48 & 46 \\
        [.5em]
        
        Sleep 		& Base     & 55 & 47 \\
         			& Enhanced & 51 & 46 \\
        [.5em]
        
        Black holes 	& Base     & 49 & 45 \\
         			& Enhanced & 48 & 47 \\
                    
        \midrule
        
       	Total 		& 		   & 300 & 277 \\
        \bottomrule
    \end{tabular}
    \caption{Number of dialogues per topic and setting, \textit{before} and \textit{after} filtering out participants who did not chat about their assigned explanandum.}
    \label{tab:participant-statistics}
\end{table}

\section{Extended Results}
\label{apdx:extended-results}

\subsection{Pre-study }
\label{apdx:pre-study}
To evaluate the participants' understanding of the task and questionnaires and to detect potential issues, we conducted a pre-study with 28 participants. We instructed the participants to interact with the LLM, explicitly referring to it as a \textit{chatbot}, and ask questions about the specific topic. We observed that their interactions differed from how they might naturally communicate with a human. One participant in our pre-study treated the LLM's answers as unchangeable, rather than telling the LLM to avoid long answers. Another participant blamed themselves for not knowing about a topic afterwards, instead of considering that the LLM could also explain poorly. Based on these findings, we decided against explicitly priming participants about who they were interacting with for our final study.

We also considered pretending that the LLM is a human to reduce barriers to natural human-to-human interaction. However, \citet{Gnewuch2018responseDelay} found that it would require significant additional effort, such as artificially slowing the response time to match human typing speed, to convince participants that they were interacting with a human. Since this is beyond the scope of this paper, we did not implement such methods.

\subsection{Study Statistics}
\label{apdx:study-statistics}

Table~\ref{tab:participant-statistics} shows the number of participants before and after filtering, across all topics and LLM settings. We exclude dialogues of participants that did not follow the provided task instructions, such as chatting about the wrong explanandum.

\subsection{Dialogue Statistics}
\label{apdx:dialogue-statistics}

Table~\ref{tab:chat-statistics-topics} shows statistics of all dialogues in a respective LLM setting and topic combination.

\begin{table*}
    \small
    \centering
    
    \renewcommand{\arraystretch}{0.9}
    \setlength{\tabcolsep}{6.5pt}
    
    \begin{tabular}{llrrrrr}
        \toprule
         & & & \multicolumn{2}{c}{\bf Explainee} & \multicolumn{2}{c}{\bf Explainer} \\
        \cmidrule(l@{3pt}r@{3pt}){4-5}\cmidrule(l@{3pt}r@{3pt}){6-7}
         
         & & \multicolumn{1}{c}{\bf Duration} & \textbf{\# Queries} & \textbf{Processing Time} & \textbf{\# Sentences} & \textbf{\# Words/sentence} \\

        \midrule
        
        Quarto & Base & \textsuperscript{$\dagger$}11:49 $\pm$ 216s & 9.3 $\pm$ 4.4 & 01:36 $\pm$ 64s & \textsuperscript{$\dagger$}14.1 $\pm$ 4.2 & \textsuperscript{$\dagger$}15.8 $\pm$ 2.0 \\
        & Enhanced & \textsuperscript{$\dagger$}13:07 $\pm$ 199s & 10.5 $\pm$ 4.7 & 01:27 $\pm$ 40s & \textsuperscript{$\dagger$}10.4 $\pm$ 3.7 & \textsuperscript{$\dagger$}17.3 $\pm$ 2.5 \\
        [.5em]
        
        Sleep & Base & \textsuperscript{$\dagger$}12:05 $\pm$ 221s & \textsuperscript{$\dagger$}6.9 $\pm$ 2.6 & 02:07 $\pm$ 73s & \textsuperscript{$\dagger$}19.3 $\pm$ 4.8 & \textsuperscript{$\dagger$}16.2 $\pm$ 3.1 \\
        & Enhanced & \textsuperscript{$\dagger$}13:39 $\pm$ 161s & \textsuperscript{$\dagger$}9.4 $\pm$ 3.5 & 01:47 $\pm$ 69s & \textsuperscript{$\dagger$}12.7 $\pm$ 3.6 & \textsuperscript{$\dagger$}17.3 $\pm$ 2.0 \\
        [.5em]
        
        Black holes & Base & 13:27 $\pm$ 197s & 8.6 $\pm$ 3.9 & 01:45 $\pm$ 49s & \textsuperscript{$\dagger$}20.1 $\pm$ 4.8 & 18.5 $\pm$ 1.6 \\
        & Enhanced & 13:29 $\pm$ 160s & 8.1 $\pm$ 2.2 & 01:46 $\pm$ 36s & \textsuperscript{$\dagger$}13.0 $\pm$ 2.8 & 18.8 $\pm$ 1.4 \\
        
        \bottomrule
    \end{tabular}
    
    \caption{Dialogue statistics per topic and LLM setting showing the \textit{duration} of the interaction between the participants (explainee) and the LLM (explainer) in minutes, the \textit{number of queries} send by the participants, the \textit{processing time} that the participants needed to respond to an LLM answer, as well as the \textit{number of sentences} and \textit{number of words per sentence} per LLM response. All numbers are averaged over all dialogues of a topic in the respective LLM setting. Significant differences between the two settings are marked with \textsuperscript{$\dagger$} ($p < 0.05$).}
    \label{tab:chat-statistics-topics}
\end{table*}

\subsection{Dialogue Act and Explanation Move Predictions}
\label{apdx:turn-level-prediction}

\begin{table*}
    \centering
    \small
    
    \renewcommand{\arraystretch}{0.9}
    \setlength{\tabcolsep}{6pt}
    
    \begin{tabular}{lrrrrrrrrr}
        \toprule
        &\multicolumn{3}{c}{\bf Explanation Moves} & \multicolumn{3}{c}{\bf Dialogue Acts} & \multicolumn{3}{c}{\textbf{Topics}} \\
        \cmidrule(l@{2pt}r@{2pt}){2-4}   \cmidrule(l@{2pt}r@{2pt}){5-7}      \cmidrule(l@{2pt}r@{2pt}){8-10}  
        
        \textbf{Model} & \bf ELI-5 & \bf 5-Levels & \bf Overall & \bf ELI-5 & \bf 5-Levels & \bf Overall & \bf ELI-5 & \bf 5-Levels & \bf Overall \\
        \midrule
        
        Ours & 0.37 & 0.38 & 0.41 & 0.38 & 0.47 & 0.48 & 0.38 & 0.56 & 0.50 \\
        \citet{alshomary:2024} & 0.35 & 0.35 & 0.39 & 0.39 & 0.48 & 0.48 & 0.40 & 0.53 & 0.50 \\

        \bottomrule
    \end{tabular}
    
    \caption{Macro F$_1$-score results in 5-fold cross validation of our classifier on the turn-level prediction of explanation moves, dialogue acts, and topics, compared to the original results reported by \citet{alshomary:2024}. Results of both approaches are produced by models trained on both corpora combined, \textit{ELI-5} and \textit{5-Levels}, evaluated on the separate test sets and the combined test set.}
    \label{tab:longformer-results}
\end{table*}

To investigate the co-constructive behavior of the LLM, we re-train a dialogue act and explanation move annotation model originally presented by \citet{alshomary:2024}. Details on the training can be found in Section~\ref{apdx:turn-level-prediction-setup}. Below, we present the results of our model and the mean proportions of the annotated dialogue acts and explanation moves for the LLM turns. In Section~\ref{sec:co-constructive-behavior-results}, the proportions of the participants' turns are discussed.\footnote{The models that we used to annotate the turns of our dialogues can be found under \url{https://huggingface.co/webis/sigdial25-co-constructive-llms}}

\paragraph{Results of Our Re-trained Model}

Table~\ref{tab:longformer-results} shows the macro F$_1$-score results of a 5-fold cross validation for annotating dialogue acts, explanation moves, and the topic with our re-trained model of the approach presented in \citet{alshomary:2024}. In contrast to \citet{alshomary:2024}, we only report results for training on both datasets, ELI-5 and 5-Levels, as this setting was reported to perform best. Our re-trained model shows comparable performance to the results reported by \citet{alshomary:2024}.

\paragraph{Results of the annotated LLM turns}

\begin{figure}
    \centering
    \includegraphics[scale=1.0]{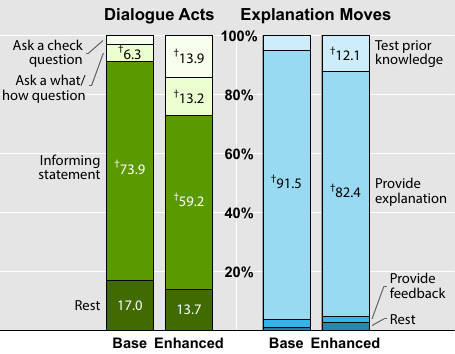}
    \caption{Proportions of annotated dialogue acts and explanation moves~\cite{wachsmuth2022wired, alshomary:2024} for the \emph{LLMs'} turns, normalized per dialogue of the \emph{base} and \emph{enhanced} settings, respectively. \emph{Rest} denotes the sum for all labels that have a proportion smaller than 2\% or that are too unspecific. Significant differences between the two settings are marked with \textsuperscript{$\dagger$} ($p < 0.05$). }
    \label{fig:turn-prediction-explainer}
\end{figure}

The mean proportions of the annotated dialogue acts and explanation moves for the LLM turns of our study dialogues are shown in Figure~\ref{fig:turn-prediction-explainer}. The label \textit{Rest} is the sum of the proportions of the labels that have a proportion smaller than 2\% or that are unspecific.

\subsection{Qualitative Analysis}
\label{apdx:qualitative-analysis}

For the qualitative analysis, we aim to select extreme cases that highlight relevant aspects. To do this, we filter the 25\% best and worst participants in terms of their understanding and co-constructiveness of the LLM. Specifically, we filter according to the increase in subjective comprehension and the level of objective comprehension after the interaction, assessed by our questionnaires. We further filter the remaining dialogues based on the potential for relevant insights regarding the co-constructive behavior of the LLM, assessed by the participants through the co-constructiveness questionnaire (see Figure~\ref{fig:co-construction-questionnaire}). This filtering results in nine dialogues that were analyzed qualitatively.

Table~\ref{tab:qualitative-examples} shows excerpts from the nine dialogues that are referenced in Section~\ref{sec:quali-analysis}, highlighting the results of our qualitative analysis.

\newcolumntype{Y}{>{\raggedright\arraybackslash}X}

\begin{table*}
    \fontsize{8.5}{10}\selectfont
    \renewcommand{\arraystretch}{0.9}
    \centering
    \setlength{\tabcolsep}{3pt}

    \begin{tabularx}{\linewidth}{lp{2cm}YX}
        \toprule

        \textbf{PID} & \textbf{Topic/ Setting} & \textbf{Dialogue Excerpt} & \textbf{Observation} \\

        \midrule

        943a & Quarto/\newline Base & \parbox{\hsize}{\texttt{0:03:39 EX:} \\\text{[\ldots]} Would you like to know more about strategies for playing Quarto, or do you have any other questions about the game? \\ \texttt{0:04:43 EE:}\\ So, the players don't hold any pieces, they play them the instant they receive them? \text{[\ldots]} \\\texttt{0:04:50 EX:} \\\text{[\ldots]}  Would you like to know more about common strategies or tactics in Quarto? \\ \texttt{0:06:32 EE:} \\ so there are 4 square, 4 round, 4 dark, 4 light 4 tall, 4 short, 4 with a hole and 4 without a hole?} & A co-constructive interaction requires effort; the explainee seems to deliberately not answer the questions, but rather forces the explainer to answer their questions. \\

        \midrule

        4be6 & Black holes/\newline Base & \parbox{\hsize}{\texttt{0:02:09 EE:} \\ I want to know what black hole is all about, note I have never heard of it before \\ \text{[}Receiving a long response\text{]} \\  \texttt{0:05:41 EE:} \\ Yeah, this is a lot can we take it bit by bit} & The explainee requests shorter explanations, with more reasonable amounts of information to digest. \\

        \midrule

        4be6 & Black holes/\newline Base & \parbox{\hsize}{\texttt{0:09:31 EX:} \\\text{[\ldots]} Now, would you like to know: \\ A) What happens when something gets too close to a black hole? \\ B) How big can black holes get? \\ C) What is the difference between a black hole and a neutron star? \\ D) Something else (please specify)? \\  \texttt{0:13:16 EE:} \\ how big can a black hole get} & Multiple-choice options from explainer, allowing the explainee to choose an obvious \textit{path of least resistence}. \\

        \midrule

        1570 & Quarto/\newline Enhanced & \parbox{\hsize}{\texttt{0:07:02 EX:} \\ \text{[\ldots]} \\ Now, going back to your previous question, can you think of a scenario where a player might want to place a piece that doesn't immediately seem beneficial to them, but might actually be a good strategic move? \\ \texttt{0:10:23 EE:} \\ I have another question, excuse me, so the selector wins by forcing the placer to place pieces where they don't want to place them, as in they're forced to play it in certain areas? } & The explainee must shift the interaction, doing extra work to suspend the explainer's question and initiate a new thread.  \\

        \midrule

        43b6 & Black holes/\newline Enhanced & \parbox{\hsize}{\texttt{0:05:54 EX:} \\ \text{[\ldots]}  Now, let's consider the environment around a supermassive black hole. What do you think happens to the stars and other objects that get too close to the event horizon? Are they slowly pulled in, or is there a more dramatic fate that awaits them? \\  \texttt{0:09:10 EE:} \\ I believe that the force of attraction intrinsic to the existence of the black hole will lead to the disintegration of these stars and consequently the incorporation of their matter into the disk surrounding the black hole.}  & Both parties co-shape the interaction. Much effort is required from the explainee for this to happen. \\

        \midrule

        2417 & Sleep/\newline Enhanced & \parbox{\hsize}{\texttt{0:03:38 EE:} \\ I often find myself feeling tired throughout the day, but i am still getting the 7-9 hours of sleep that you said. \\ \texttt{0:03:45 EX:} \\ Feeling tired despite getting 7-9 hours of sleep can be a bit puzzling. There are a few possible explanations for this. [\ldots]} & The participants contributes personal experiences to the interaction, enabling the conversation to shift away from purely monological explanations. \\

        \bottomrule
    \end{tabularx}

    \caption{Relevant examples of the qualitative analysis, showing excerpts from selected dialogues between the explainer (\textit{EX}) and the explainee (\textit{EE}) that highlight interesting interaction patterns. The Participant ID (\textit{PID}) column shows the first four characters of the unique participant identifier. For more details, refer to Section~\ref{sec:quali-analysis}.}
    \label{tab:qualitative-examples}
\end{table*}

\end{document}